\documentclass[10pt,twocolumn,letterpaper]{article}

\usepackage{wacv}
\usepackage{times}
\usepackage{epsfig}
\usepackage{graphicx}
\usepackage{amsmath}
\usepackage{amssymb}

\usepackage{inputenc}
\usepackage[T1]{fontenc}
\usepackage{verbatim}
\usepackage{multirow}
\usepackage[algoruled,boxed]{algorithm2e}
\usepackage{xcolor}
\usepackage{colortbl}
\usepackage{subcaption}
\usepackage{float}
\usepackage{enumitem}
\setlist{nolistsep}
\usepackage{amssymb}
\usepackage{wrapfig}

\definecolor{Gray0}{gray}{0.90}
\definecolor{Gray1}{gray}{0.80}

\usepackage[pagebackref=true,breaklinks=true,letterpaper=true,colorlinks,bookmarks=false]{hyperref}

\wacvfinalcopy 

\def\wacvPaperID{****} 

\ifwacvfinal\fi
\begin{document}

\title{Relational Mimic for Visual Adversarial Imitation Learning}



\author{%
Lionel Blond\'e\thanks{Work done during an internship at Apple.} \\
University of Geneva, Switzerland \\
{\tt\small lionel.blonde@etu.unige.ch}
\and
\hspace{-0.5cm} Yichuan Charlie Tang \hspace{1.0cm} Jian Zhang \hspace{1cm} Russ Webb \\
\hspace{1.2cm} Apple Inc. \\
{\tt\small \{yichuan\_tang, jianz, rwebb\}@apple.com}
}

\maketitle
\ifwacvfinal\thispagestyle{empty}\fi

\begin{abstract}
    In this work, we introduce a new method for imitation learning from video demonstrations.
    Our method, \textit{Relational Mimic} (\textsc{RM}),
    improves on previous visual imitation learning methods by combining
    generative adversarial networks and relational learning.
    \textsc{RM} is flexible and can be used in conjunction with other recent advances
    in generative adversarial imitation learning \cite{Ho2016-bv}
    to better address the need for more
    robust and sample-efficient approaches.
    In addition, we introduce a new neural network architecture
    that improves upon the previous state-of-the-art in reinforcement learning and illustrate
    how increasing the relational reasoning capabilities of the agent enables the latter
    to achieve increasingly higher performance in a challenging locomotion task with pixel inputs.
    Finally, we study the effects and contributions of relational learning in policy evaluation,
    policy improvement and reward learning through ablation studies.
\end{abstract}

\section{Introduction}

    Reinforcement learning (RL) \cite{Sutton1998-ow}
    has received attention
    for training intelligent agents capable of solving sequential decision-making tasks under uncertainty.
    While RL has been hindered by the lack of strong function approximators,
    the advances in model expressiveness enabled by deep neural networks have enabled deep reinforcement learning
    to solve increasingly challenging tasks \cite{Mnih2015-iy,Silver2016-my,OpenAI2018-sm}.
    Despite alleviating the burden of hand-crafting task-relevant features,
    deep RL is constrained by the need for reward shaping \cite{Ng1999-lv},
    that is, designing a reward signal that will guide the learning agent to solve the end-task.

    Instead of relying on reward signals
    requiring considerable engineering efforts to align \cite{Leike2018-it} with the desired goal,
    we use imitation learning (IL) \cite{Bagnell2015-ni}
    in which the agent is provided with expert demonstrations before the training procedure starts.
    The agent does not receive any external reward signal while it interacts with its environment.
    The behavior that emerges from the mimicking agent should resemble the behavior demonstrated by the expert.
    Demonstrations have successfully helped mitigate side effects due to
    reward hand-crafting, commonly referred to as \textit{reward hacking} \cite{Hadfield-Menell2017-or},
    and induce risk-averse, safer behaviors \cite{Amodei2016-tb,Lacotte2019-kx,Leike2017-nc}.

    When the demonstrated trajectories contain both the states visited by the expert
    and the controls performed by the expert in each state,
    the imitation learning task can be framed as a supervised learning task,
    where the agent's objective consists of learning the mapping between states and controls
    (commonly denoted as actions in the RL framework).
    This supervised approach is referred to as behavioral cloning (BC),
    and has enabled advances
    in autonomous driving \cite{Pomerleau1989-nh,Pomerleau1990-lm,Bojarski2016-uz}
    and robotics \cite{Ratliff2007-fc,Argall2009-ga,Torabi2018-fg}.
    However, BC remains extremely brittle in the absence of abundant data.
    In particular, the cloning agent can only recover from past mistakes
    if corrective behavior appears in the provided demonstrations
    due to compounding errors.
    This phenomenon, known as \textit{covariate shift} \cite{Ross2010-eb,Ross2011-dn},
    showcases the fragility of supervised learning approaches in interactive,
    dynamically-entangled, sequential problems.

    In contrast to BC, Apprenticeship learning (AL)
    \cite{Abbeel2004-rb} tackles IL problems without
    attempting to map states to actions in a supervised learning fashion.
    AL tries to first recover the reward signal which explains the behavior observed in the demonstrations,
    an approach called inverse reinforcement learning (IRL) \cite{Ng2000-qd,Bagnell2015-ni},
    and subsequently uses the learned reward to train the agent by regular RL.
    Assuming the recovered reward is the reinforcement signal that was optimized by the expert demonstrator,
    learning a policy by RL from this signal will yield a policy mimicking the demonstrated behavior.
    While training models with data interactively collected from the environment
    mitigates the compounding of errors,
    solving an RL problem every iteration (for every new reward update) is expensive,
    and recovering the expert's exact reward is an ill-posed problem \cite{Ziebart2008-fe},
    often requiring various relaxations \cite{Neu2012-fw,Syed2008-zo,Syed2008-su,Syed2010-ke,Ho2016-xn}.

    Generative adversarial imitation learning (GAIL) \cite{Ho2016-bv}
    addresses these limitations
    by jointly
    learning a similarity metric with a generative adversarial network (GAN) \cite{Goodfellow2014-yk}
    and
    optimizing a policy by RL using the learned similarity metric as a surrogate reward.
    In contrast to AL,
    GAIL does not try to recover the reward signal assumed to have been optimized by the expert
    when demonstrating the target behavior.
    Instead, GAIL learns a surrogate signal that, when learned jointly with the policy which uses it as a reward,
    yields a robust, high-performance imitation policy.
    Interestingly, in environments that are close to deterministic
    (which applies to every current benchmark environment),
    GAIL still performs well when provided with expert demonstrations
    containing only the states visited by the expert,
    without the actions that led the expert from one state to the next \cite{Torabi2018-nb}.
    This is important because video sharing platforms provide numerous instances of state-only demonstrations.
    Being able to leverage these,
    by designing imitation methods which use videos as input,
    is an important imitation learning milestone.
    In this work, we introduce \textsc{RM}, an video imitation learning method that builds on GAIL.

\begin{figure}
    \centering
    \begin{subfigure}[t]{.40\linewidth}
    \includegraphics[width=\linewidth]{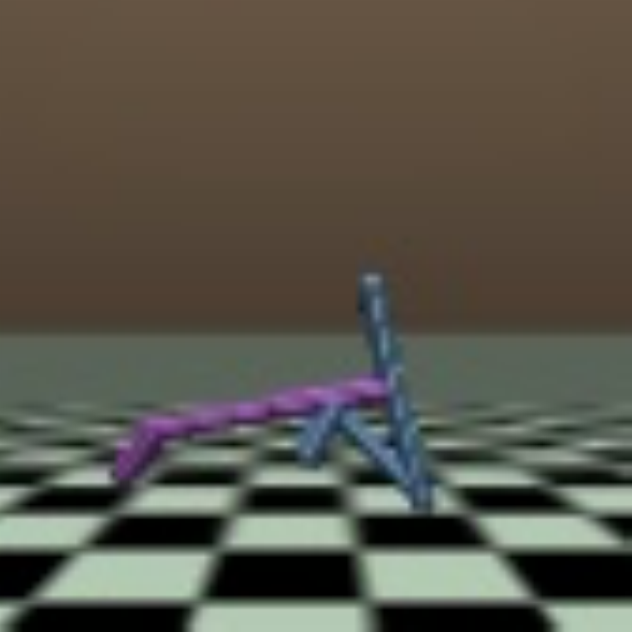}
    \caption{}
    \label{fig:visualstate}
    \end{subfigure}
    \begin{subfigure}[t]{.56\linewidth}
    \includegraphics[width=\linewidth]{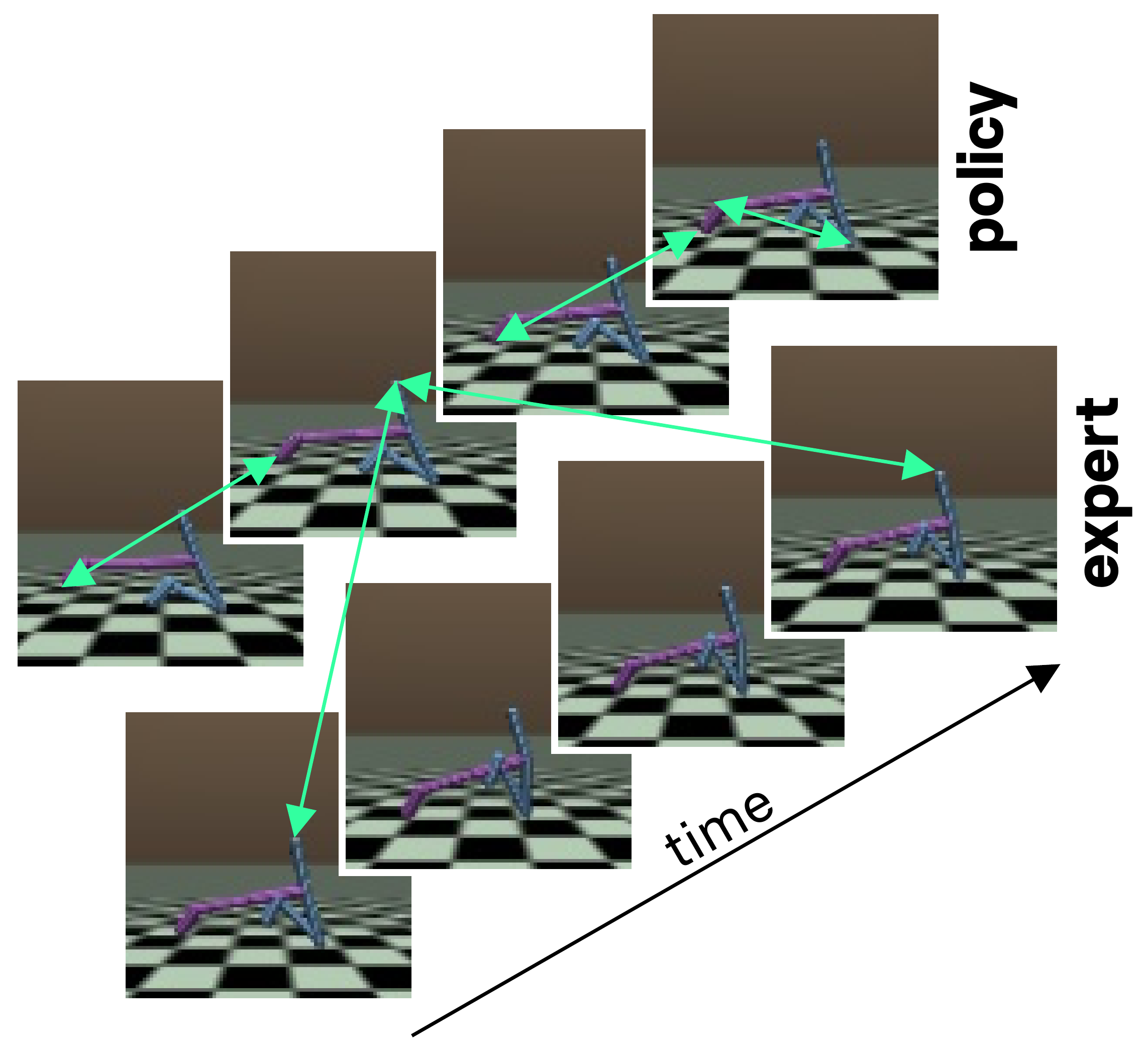}
    \caption{}
    \label{fig:seq}
    \end{subfigure}
    \caption{(a) Example frame for learning locomotion
    policies from pixel input in the Walker2d-v3 environment
    from the \textsc{MuJoCo} benchmark.
    (b) Relational learning capabilities of our approach,
    with examples of possible relationships depicted by double-ended green arrows.}
    \end{figure}

    We focus on locomotion tasks from the \textsc{MuJoCo} \cite{Todorov2012-gc}
    suite of continuous control environments \cite{Brockman2016-un}.
    In addition to control sensitivity inherent to continuous actions spaces,
    locomotion tasks require agents to maintain dynamic equilibria while solving the task.
    Learning to preserve balance is not an universal skill and depends both on the agent's body configuration
    and the goal task.
    When the proprioceptive state representation of the agent is available (joint angles, positions and velocities),
    introducing a structural bias by explicitly modeling the morphology of the agent
    to learn the relationships between joints as edges in a graph
    can yield significant gains in performance and robustness \cite{Wang2018-ot}.
    Rather than being provided with proprioceptive states,
    our agents are solely provided with high-dimensional visual state representations,
    preventing us from modelling the agent's structure explicitly.
    In this work,
    we introduce a self-attention mechanism in the convolutional perception stack of our agents,
    drawing inspiration from relational modules \cite{Wang2018-ti,Santoro2017-nb,Vaswani2017-lk}
    to give them the ability to perform
    visual relational reasoning from raw visual state representations.
    By working over sequences of consecutive frames,
    our agents are able to recover motion information
    and consequently can draw long-range relationships across time and space
    (\textsc{Figure} \ref{fig:seq}).
    As we show in \textsc{Section} \ref{resultsection}, \textsc{RM}
    achieves state-of-the-art performance
    in state-only visual imitation
    in complex locomotion environments.

\section{Background}

    We model the stateful environment, $\mathcal{E}$, with a \textit{Markov decision process} (MDP),
    formally represented by the tuple
    $(\mathcal{S}, \mathcal{A}, \rho_0, P, r, \gamma)$,
    containing:
    the state space, $\mathcal{S}$,
    the action space, $\mathcal{A}$,
    the initial state density, $\rho_0$,
    and transition distribution with conditional density, $P(s_{t+1}|s_t, a_t)$,
    jointly defining the world dynamics,
    the reward function, $r: \mathcal{S} \times \mathcal{A} \to \mathbb{R}$,
    and the discount factor, $\gamma$, as traditionally seen in infinite-horizon MDPs.
    Since we work in the episodic setting, we assume that every agent-generated trajectory
    ends in an absorbing state at time horizon $T$
    which triggers episode termination by zeroing out $\gamma$
    once reached.
    The environment is unknown to the agent and can only be queried via interaction.

    The agent follows its policy, $\pi_\theta$,
    modeled via a neural network parametrized by $\theta$,
    to sequentially determine which decision (action), $a_t \in \mathcal{A}$, to make
    in the current situation (state), $s_t \in \mathcal{S}$.
    The decision process is dictated by the conditional density $\pi_\theta(a_t|s_t)$.
    \emph{In contrast with traditional RL settings, our agent is not rewarded upon interaction
    and therefore does not benefit from external supervision towards completing the task at hand.}
    The agent however has access to a set of trajectories collected by an \textit{expert} policy,
    $\pi_e$, which demonstrates how to complete the task.
    Our objective is to train agents capable of displaying behaviors resembling the ones shown
    by the expert in the demonstrations with high fidelity and in a robust fashion.
    Since we consider the problem of imitation learning from states without the associated actions
    performed by the expert,
    we define \textit{demonstrations} as sequences of states visited by
    the expert during an episode, $\{s_0, \ldots, s_T\}$.
    Finally, we introduce the \textit{return}
    defined as the sum of future rewards collected from time $t$ to episode termination,
    $R_t^\gamma \triangleq \sum_{k=t}^{\infty} \gamma^{k-t}r(s_t, a_t)$,
    and the \textit{state value} defined as the expected return of starting from state, $s_t$
    and following policy $\pi_\theta$ thereafter,
    $V^{\pi_\theta}(s_t) \triangleq \mathbb{E}_{a_t \sim \pi_\theta(\cdot | s_t),
    s_{t+1} \sim P(\cdot | s_t, a_t),
    a_{t+1} \sim \pi_\theta(\cdot | s_{t+1}), \ldots}
    [R_t^\gamma]$.
    Since we are interested in the representations learned for policy evaluation,
    we learn $V^{\pi_\theta}$ (as opposed to simply computing it, \textit{e.g.}, by Monte-Carlo estimation).
    We model $V^{\pi_\theta}$ via a neural network $V_\phi$, parametrized by $\phi$.

\section{Relational Mimic}

    We introduce \textit{Relational Mimic} (\textsc{RM}),
    capable of learning efficient imitation policies from visual input
    in complex locomotion environments.
    To learn in complex environments,
    agents have to deal with many limbs,
    connected by complex joints (with potentially many degrees-of-freedom),
    which calls for greater coordination skills.
    Our experiments show that using relational modules in our agent is key to the success of our method.

\begin{figure}
        \center\scalebox{0.24}[0.24]{\includegraphics{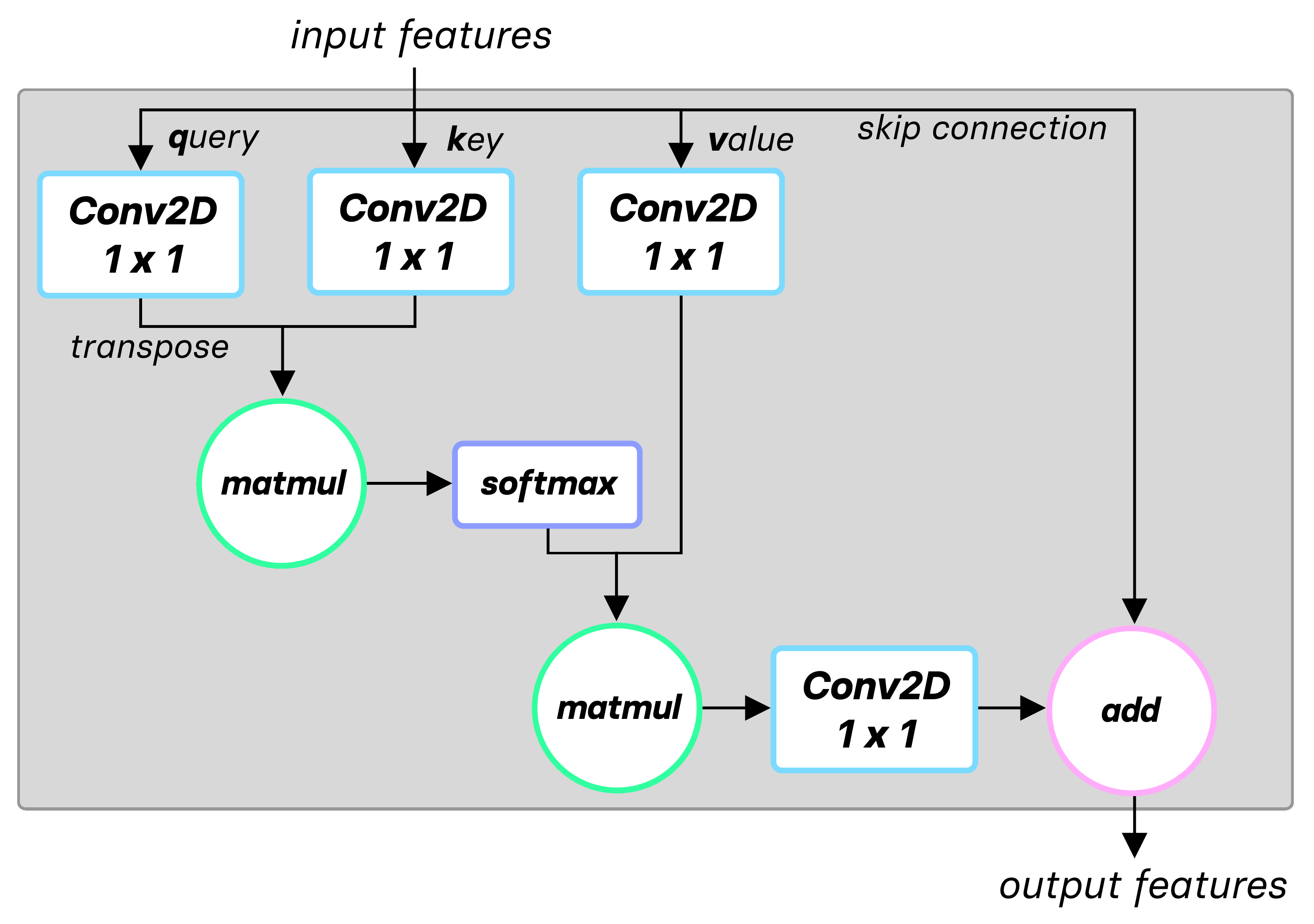}}
        \caption{Relational Block.
        \textit{``matmul''}, \textit{``softmax''}, and \textit{``add''}
        respectively denote the matrix multiplication,
        row-wise softmax,
        and addition operations.}
        \label{fig:relblock}
        \end{figure}

    We first focus on a new architecture for visual continuous control
    which will serve as a building block throughout the remainder of the paper.
    This architecture takes advantage of non-local modeling,
    which has arguably been overlooked in recent successful deep vision-purposed architectures
    and has very recently been revived in \cite{Wang2018-ti}.
    When used in a reinforcement learning scenario,
    we show that an agent
    using the proposed architecture
    not only improves on previous canonical architectures for control
    (\textsc{Figures} \ref{fig:archicomparison} and \ref{fig:rl}),
    but also that enriching the architecture with a (slightly modified)
    \textit{non-local block} \cite{Wang2018-ti}
    (\textsc{Figure} \ref{fig:relblock})
    yields even further improvements in the locomotion tasks tackled.
    For legibility purposes, we will hereafter refer to the agent built with
    the complete architecture as \textit{non-local agent}
    (\textsc{Table} \ref{fig:fullarchi}),
    and to the agent without non-local block as \textit{local agent}.
    We then introduce our main contribution,
    \textsc{RM}, which builds on recent advances in both relational reasoning and
    GANs.

    \textbf{Visual inputs.}
        As previously mentioned, our agents do not make use of proprioceptive feedback,
        which provides the agent with detailed, salient information about its current state
        (joint angles and velocities).
        Instead, our agents perceive only a \textit{visual} (pixel) representation of the state
        (see \textsc{Figure} \ref{fig:visualstate}),
        resulting from passing the proprioceptive state into a renderer
        before giving it to the agent (more details found in \textsc{Section} \ref{resultsection}
        describing the experimental setup).
        While the joint positions can be recovered with relative ease depending on the resolution of
        the rendered image,
        joint velocities are not represented in the rendered pixel state.
        In effect, the state is now partially-observable.
        However, to act optimally, the agent needs to know its current joint velocities.
        This hindrance has previously been addressed either by
        using frame-stacking or using a recurrent model.
        Agents modeled with recurrent models learn a state embedding from single input frames
        and work over sequences of such embedding to make a decision.
        With frame-stacking however, agents learn a state sequence embedding directly,
        albeit over usually shorter sequence lengths.
        Recent work reports similar performance with either approach
        \cite{Cobbe2018-kq}.
        We opt for frame-stacking to learn convolution-based
        relational embeddings over sequences of states, a key contribution of our approach.
        We denote by $k$ the number of frames stacked in the input (corresponding to back-steps),
        and define a \textit{stacked state} as the tuple
        $s_t^k \triangleq (s_{t-k+1}, \ldots, s_{t-1}, s_t)$.

    \textbf{Architecture.}
        Relational learning capabilities are given to our method by integrating
        \textit{self-attention} into the architecture.
        In \cite{Wang2018-ti}, self-attention enables the enriched model to reason in both time and space,
        by enabling the model to capture long-range dependencies both spatially within the images
        and temporally across the sequences of consecutive frames.
        Non-local operations capture global long-range dependencies
        by directly computing interactions between two positions,
        irrespective of their spatial distance and temporal delay.
        As a result,
        models augmented with self-attention mechanisms
        are naturally able to capture relationships between locations in and across frames,
        which imbues the system with the ability to perform relational reasoning
        about the objects evolving in the input.
        The architecture of our \textit{non-local agent} is described in \textsc{Table} \ref{fig:fullarchi}.
        The careful use of feature-pooling (rows 3 and 7 in \textsc{Table} \ref{fig:fullarchi})
        combined with the introduction of a self-attentive relational block
        (row 5 in \textsc{Table} \ref{fig:fullarchi})
        are the key features of our approach, and are described in dedicated sub-sections.

        In recent years, it has been common practice to use a single network with two heads,
        an action head and a value head,
        to parameterize the policy and value networks respectively.
        While reducing the computational cost,
        weight sharing relies upon the assumption that the value and policy are both optimal under
        the same feature representation.
        However, \cite{Wang2016-qf} shows with saliency maps that policy and value networks
        learn different hidden representations and pay attention to different cues in the visual input state.
        Inspired by this observation, we separate the policy and value into distinct networks,
        which will also enable us to conduct an ablation study on the effect of self-attention.

        \begin{table}
        \begin{subfigure}[t]{\linewidth}
        \centering
        \begin{tabular}{|c|l|}
        \hline
        0 & \textsc{Input}, {\footnotesize size: $[84 \times 84 \times (\text{\#colors} \times k)]$} \\
        \hline\hline
        1 & Standardize input: {\footnotesize $x \mapsto x/255$} \\ \hline
        2 & \textsc{Conv2D}, {\footnotesize $[7 \times 7]$, $[2 \times 2]$, $[3 \times 3]$, ReLU \textit{post}} \\ \hline
        \rowcolor{Gray0}
        3 & \textsc{MaxPool3D}, {\footnotesize $[3 \times 3 \times 3]$, $[2 \times 2 \times 2]$, $[1 \times 1 \times 1]$} \\ \hline
        4 & \textsc{Residual Block} \\ \hline
        \rowcolor{Gray1}
        5 & \textsc{Relational Block} \\ \hline
        6 & \textsc{Residual Block}, {\footnotesize ReLU \textit{post}} \\ \hline
        \rowcolor{Gray0}
        7 & \textsc{MaxPool3D}, {\footnotesize $[3 \times 1 \times 1]$, $[2 \times 1 \times 1]$, $[1 \times 0 \times 0]$} \\ \hline
        8 & \textsc{Residual Block} \\ \hline
        9 & \textsc{Residual Block}, {\footnotesize ReLU \textit{post}} \\ \hline
        10 & \textsc{FullyConnected}, {\footnotesize $256$} \\ \hline
        \end{tabular}
        \caption{Architecture of our \textsc{Non-Local Agent}.
        Removing the fifth row yields the \textsc{Local Agent}.}
        \label{fig:fullarchi}
        \end{subfigure}
        ~
        \vspace{0.1cm}
        \begin{subfigure}[t]{\linewidth}
        \centering
        \begin{tabular}{|c|l|}
        \hline
        0 & \textsc{Input}, {\footnotesize inner feature maps} \\
        \hline\hline
        1 & \textsc{Conv2D}, {\footnotesize $[1 \times 1]$, $[1 \times 1]$, $[0 \times 0]$, ReLU \textit{pre}} \\ \hline
        2 & \textsc{Conv2D}, {\footnotesize $[3 \times 3]$, $[1 \times 1]$, $[0 \times 0]$, ReLU \textit{pre}} \\ \hline
        3 & \textsc{Conv2D}, {\footnotesize $[1 \times 1]$, $[1 \times 1]$, $[0 \times 0]$, ReLU \textit{pre}} \\ \hline
        4 & \textsc{SkipConnection}, add input to the output \\ \hline
        \end{tabular}
        \caption{Residual Block \cite{He2015-zv} with ReLU pre-activations.}
        \end{subfigure}

        \caption{\textsc{Non-Local Agent} architecture description. Indices in the left column are an indication of depth
        (the higher the index, the deeper the associated layer is in the architecture).
        The mention "\textit{pre}" indicates that the non-linearity is applied right before the assigned row (pre-activation).
        Analogously, "\textit{post}" designates post-activations.
        The number of channels used has been fine-tuned such that the resulting model is comparable with the baselines in
        terms of convolutional stack depth, number of parameters and forward pass computational cost.
        }
        \end{table}

        \begin{table*}
        \centering
        \begin{tabular}{l|c|c|c}
        \multirow{3}{*}{\textsc{Architecture}} & \multicolumn{3}{c}{\textit{number of ...}} \\
        & \textsc{Parameters} & \textsc{Flops} & \textsc{Convolutional Layers} \\
        \hline\hline
        \textsc{Nature} \cite{Mnih2015-iy} & 1.276M & 16.62M & 3 \\ \hline
        \textsc{Large Impala} \cite{Espeholt2018-gl} & 0.5002M & 84.52M & 15 \\ \hline
        \rowcolor{Gray0}
        \textsc{Local Agent} (ours) & 0.4575M & 13.71M & 13 \\ \hline
        \rowcolor{Gray1}
        \textsc{Non-Local Agent} (ours) & 0.4577M & 13.84M & 15\\ \hline
        \end{tabular}
        \caption{Architecture comparison between the introduced architectures
        and notable convolutional baselines from the Reinforcement Learning literature.
        Criteria are, ordered from left to right,
        1) the total number of model parameters including the fully connected layers
        following the convolutional stack,
        2) the computational cost of one forward pass through the model, expressed in flops,
        and
        3) the depth of the perception stack,
        corresponding to the maximum number of consecutive convolutional layers used in the network.
        }
        \label{fig:archicomparison}
        \end{table*}

    \textbf{Feature pooling.}
        In locomotion tasks, uncertainties in the visual state representation
        can prevent the agent from infering its position accurately.
        This phenomenon is exarcerbated when the agent also has to infer its velocity from
        sequences of visual states.
        However, both position (spatial) and velocity (temporal)
        are crucial to predict sensible continuous actions.
        In order to propagate this spatial and temporal information
        as the visual input sequence is fed through the layers,
        our architecture only involves one \textit{spatial} pooling layer
        (row 3 in \textsc{Table} \ref{fig:fullarchi}), as opposed to three in
        the \textit{large Impala} network \cite{Espeholt2018-gl}.
        Note, the \textit{non-local} and \textit{large Impala} agents both have
        15-layer deep convolutional stacks.
        In order to remain competitive with the architectures in
        \textsc{Table} \ref{fig:archicomparison} in terms of number of parameters and computational cost,
        we use two \textit{feature pooling} layers, at different depth levels
        (row 3 and 7 in \textsc{Table} \ref{fig:fullarchi}, where row 3 performs both spatial and feature pooling).

    \textbf{Relational block.}
        Our relational block, shown in \textsc{Figure} \ref{fig:relblock},
        is based on \cite{Wang2018-ti}
        which implements the non-local mean operation \cite{Buades2005-gv} as follows:
        \begin{align}
        u_i \mapsto \frac{1}{N(u_i)}
        \sum_{j=0}^m
        \sigma(u_i, u_j) \,
        v(u_j)
        \;\text{.}
        \end{align}
        We seek a pairwise similarity measure, $\sigma(u_i, u_j)$,
        that considers the relationships between every pair of features
        $(u_i, u_j)$  with $0 \leq i, j \leq m$.
        Instead of comparing $u_i$ to $u_j$ directly,
        we compare embeddings derived from these features via respective $1 \times 1$ convolutional layers,
        $q(u_i) \triangleq W_q u_i$ and $k(u_j) \triangleq W_k u_j$.
        We follow non-local modeling \cite{Buades2005-gv}, and
        similarly to \cite{Wang2018-ti},
        use an exponentiated dot-product similarity
        (Gaussian radial basis function)
        between embeddings,
        $\sigma(u_i, u_j) \triangleq \exp [q(u_i)^T k(u_j)]$.
        By formalizing the normalization factor as
        $N(u_i) \triangleq \sum_{j=0}^m \sigma(u_i, u_j)$,
        we use a \textit{softmax} operation
        and implement the self-attention formulation proposed in \cite{Vaswani2017-lk}:
        \begin{align}
        u_i \mapsto
        \sum_{j=0}^m
        \operatorname{softmax}
        \left ( q(u_i)^T k(u_j) \right ) \,
        v(u_j)
        \end{align}
        where $v: x \mapsto W_v x$ denotes a position-wise linear embedding on $u_j$ computed via $1 \times 1$ convolutions.
        We lastly introduce embedding $e: x \mapsto W_e x$ (similar to $v$) on the output of the non-local mean operation
        and add a residual connection \cite{He2015-zv}
        to encourage the module to focus on local features before establishing distant dependencies:
        \begin{align}
        u_i \mapsto
        e \left (
        \sum_{j=0}^m
        \operatorname{softmax}
        \left ( q(u_i)^T k(u_j) \right ) \,
        v(u_j)
        \right )
        + u_i
        \;\text{.}
        \end{align}
        The embeddings $q$ and $k$ involved in the pairwise function $\sigma$
        use half the number channels used in $u$.
        We did not use batch normalization \cite{Ioffe2015-ls} in the relational block
        as early experiments showed it reduced performance when used in the policy network.
        This degradation was not as significant when used in the value network.
        Additionally, preliminary experiments showed a slight decrease in performance
        when using the dot-product for $\sigma$.
        To reiterate, the relational block
        combined with frame stacking enables the model to consider relationships between
        entities across space and time, while using computationally-efficient 2D convolutions.

    \textbf{Reward learning.}
        \begin{table}
        \centering
        \begin{tabular}{|c|l|}
        \hline
        0 & \textsc{Input}, {\footnotesize size: $[84 \times 84 \times (\text{\#colors} \times k)]$} \\
        \hline\hline
        1 & Standardize input: {\footnotesize $x \mapsto x/255$} \\ \hline
        2 & \textsc{Conv2D}, {\footnotesize $[4 \times 4]$, $[2 \times 2]$, $[1 \times 1]$, LReLU \textit{post}} \\ \hline
        3 & \textsc{Conv2D}, {\footnotesize $[4 \times 4]$, $[2 \times 2]$, $[1 \times 1]$, LReLU \textit{post}} \\ \hline
        4 & \textsc{Conv2D}, {\footnotesize $[4 \times 4]$, $[2 \times 2]$, $[1 \times 1]$, LReLU \textit{post}} \\ \hline
        \rowcolor{Gray1}
        5 & \textsc{Relational Block} \\ \hline
        6 & \textsc{Conv2D}, {\footnotesize $[4 \times 4]$, $[2 \times 2]$, $[1 \times 1]$, LReLU \textit{post}} \\ \hline
        \rowcolor{Gray1}
        7 & \textsc{Relational Block} \\ \hline
        8 & \textsc{Conv2D}, {\footnotesize $[4 \times 4]$, $[2 \times 2]$, $[1 \times 1]$, LReLU \textit{post}} \\ \hline
        9 & \textsc{FullyConnected}, {\footnotesize $256$} \\ \hline
        \end{tabular}
        \caption{Reward network architecture used in \textsc{RM}.
        \textit{LReLU} designates \textit{leaky ReLU} activations \cite{Maas2013-dw}
        with a \textit{leak} coefficient
        (slope for $x<0$)
        of $0.1$.
        }
        \label{fig:discarchi}
        \end{table}
        Generative adversarial imitation learning (GAIL) \cite{Ho2016-bv}
        proposes a new approach to apprenticeship learning
        \cite{Abbeel2004-rb,Neu2012-fw,Syed2008-zo,Syed2008-su,Syed2010-ke,Ho2016-xn}
        by deriving a surrogate reward instead of trying to recover the true reward
        explaining of the expert behavior.
        The GAIL framework involves a generative adversarial network (GAN) \cite{Goodfellow2014-yk},
        opposing two networks, a \textit{generator} and a \textit{discriminator},
        in a two-player zero-sum game.
        The role of generator is played by the policy, $\pi_\theta$,
        which attempts to pick expert-like actions to arrive at expert-like states.
        The role of discriminator is played by a binary classifier with parameter vector $\omega$ denoted as $D_\omega$.
        The discriminator observes samples from both expert and policy and tries to tell them apart.
        In GAIL, the success of a policy, $\pi_\theta$, is measured by its ability to fool $D_\omega$
        into outputting that state-action pairs collected by $\pi_\theta$ were collected by $\pi_e$.
        This setup can be formulated as a minimax problem
        $\min_\theta \max_\omega \, [V(\theta, \omega)]$,
        where
        \begin{align}
        V(\theta, \omega) \triangleq
        \mathbb{E}_{\pi_\theta}[\log(1 - D_\omega(s, a))] + \mathbb{E}_{\pi_e}[\log D_\omega(s, a)]
        \;\text{.}
        \end{align}
        Observing that $D_\omega(s,a)$ measures how much a state-action pair generated by the
        policy resembles the expert's state-action pairs,
        \textsc{RM} builds on GAIL by making use of the learned $D_\omega(s,a)$ to craft a surrogate reward
        and training $\pi_\theta$ to maximize this surrogate reward by policy optimization.
        The introduced $\omega-$parameterized reward signal is trained by minimizing the regularized
        cross-entropy loss $\ell_\text{CE}$:
        \begin{align}
        \min_\omega \, \ell_\text{CE}(\omega) \triangleq
        \min_\omega \, [- V(\theta, \omega)
        + \nu \operatorname{GP}(\omega)]
        \end{align}
        where $\operatorname{GP}(\omega)$ denotes the usual gradient penalty regularizer \cite{Gulrajani2017-mr},
        encouraging the discriminator to be near-linear and consequently near-convex.
        Originally used to mitigate destructive $\omega$ updates in Wasserstein GANs \cite{Arjovsky2017-la},
        many works have hitherto reported analogous benefits
        when used in the original JS-GANs \cite{Fedus2018-bk,Lucic2017-nz}.
        Mitigating layer-wise spectral norm in $\omega$ \cite{Miyato2018-wc} yields similar stability advantages
        and is less computationally intensive than the gradient penalty.
        We use spectral normalization \cite{Miyato2018-wc}
        and gradient penalization \cite{Gulrajani2017-mr}
        simultaneously, as advocated in \cite{Kurach2018-cs},
        to train $\omega$ in \textsc{RM}.

        We align the surrogate reward signal, $r_\omega(s, a)$,
        with the confusion of the classifier.
        If the policy manages to fool the classifier, the policy is rewarded for its prediction.
        To ensure numerical stability,
        we define the reward as:
        $
        r_\omega(s, a) \triangleq
        -\log(1 - D_\omega(s, a))
        $.
        The parameters $\omega$ are updated every iteration by descending along gradients of
        $\ell_\text{CE}(\omega)$ evaluated alternatively on
        a) the minibatch collected by $\pi_\theta$ earlier in the iteration
        and
        b) the expert demonstrations.
        The demonstrations however do not contain actions.
        Nevertheless, in environments with near-deterministic dynamics,
        action $a_t$ can be approximatively inferred from the state $s_t$ and the next $s_{t+1}$.
        Consequently, $(s_t, s_{t+1})$ constitutes a good proxy for $(s_t, a_t)$.
        Note, $s_{t+1}$ is available to the agent as soon as the control, $a_t$, is executed in $s_t$.
        Since we defined the \textit{stacked state} $s_t^k$ as the sequence of $k$ latest observations,
        $(s_{t-k+1}, \ldots, s_{t-1}, s_t)$,
        the two most recent observations in $s_{t+1}^k$ are $s_t$ and $s_{t+1}$.
        We can therefore evaluate $(s_t, s_{t+1})$ by evaluating $s_{t+1}^k$, and
        define $r_\omega$ over stacked states:
        \begin{align}
        r_\omega(s_{t+1}^k) \triangleq
        -\log(1 - D_\omega(s_{t+1}^k))
        \;\text{.}
        \end{align}
        The architecture of the $\omega$-parameterized network is depicted in \textsc{Table} \ref{fig:discarchi},
        and draws inspiration from the one in \cite{Zhang2019-eb}.
        The network features two relational blocks
        which makes it capable of performing \textit{message passing} between
        distant locations (in time, space or both) in the input.

        \textbf{\textsc{RM} training.}
        The value and policy networks are trained with the PPO algorithm \cite{Schulman2017-ou}.
        \textsc{Algorithm} \ref{fig:algo} provides more details on the procedure.

        \begin{algorithm}
        \SetKwComment{Comment}{}{}
        Initialize network parameters ($\theta$, $\phi$, $\omega$) \\
        \For{$\text{i} \in 1, \ldots, \text{i}_\text{max}$}{
            \For{$\text{c} \in 1, \ldots, \text{c}_\text{max}$}{
                Observe $s_t^k$ in environment $\mathcal{E}$,
                perform action $a_t \sim \pi_\theta(\cdot | s_t^k)$,
                observe $s_{t+1}^k$ returned by $\mathcal{E}$ \\
                Augment $(s_t^k, a_t, s_{t+1}^k)$ with reward $r_\omega(s_{t+1}^k)$
                and store
                $(s_t^k, a_t, s_{t+1}^k, r_\omega(s_{t+1}^k))$ in $\mathcal{C}$ \\
            }
            \For{$\text{t} \in 1, \ldots, \text{t}_\text{max}$}{
                \For{$\text{d} \in 1, \ldots, \text{d}_\text{max}$}{
                    Sample uniformly a minibatch $\mathcal{B}^c$ of
                    \textit{stacked} states $s^k$
                    from $\mathcal{C}$ \\
                    Sample uniformly a minibatch $\mathcal{B}_e^c$ of
                    \textit{stacked} states $s^k$
                    from the expert dataset $\tau_e$,
                    with $|\mathcal{B}^c| = |\mathcal{B}_e^c|$ \\
                    \textbf{Update reward} parameters $\omega$ with the equal mixture
                    $\mathcal{B}^c \cup \mathcal{B}_e^c$ by following the gradient:
                    $\hat{\mathbb{E}}_{\mathcal{B}^c}
                    [\nabla_\omega \log(1 - D_\omega(s^k))]
                    +
                    \hat{\mathbb{E}}_{\mathcal{B}_e^c}
                    [\nabla_\omega \log D_\omega(s^k)]
                    +
                    \nu \operatorname{GP}(\omega)$
                }
                \For{$\text{g} \in 1, \ldots, \text{g}_\text{max}$}{
                    Sample uniformly a minibatch $\mathcal{B}^g$ of
                    augmented transitions
                    from $\mathcal{C}$ \\
                    \textbf{Update policy} parameters $\theta$
                    and \textbf{value} parameters $\phi$
                    with PPO \cite{Schulman2017-ou} with the minibatch $\mathcal{B}^g$
                }
            }
            Flush $\mathcal{C}$
        }
        \caption{Relational Mimic}
        \label{fig:algo}
        \end{algorithm}

\section{Related Work}

    \textbf{Self-Attention.}
    Self-attention has allowed significant advances in neural machine translation \cite{Vaswani2017-lk}
    and has improved upon previous state-of-the-art methods,
    which often relied on recurrent models,
    in the various disciplines of Machine Learning
    \cite{Ramachandran2019-yh,Velickovic2018-dw,Bello2019-kt,Zhang2019-eb}.
    Not only does self-attention enable state-of-the-art results in sequence prediction and generation,
    but it can also be formalized as a non-local operation
    (closely related to the \textit{non-local mean} \cite{Buades2005-gv}).
    This operation is straightforward to integrate into deep convolutional architectures
    to solve video prediction tasks, alleviating the need to introduce difficult to train and
    tedious to parallelize recurrent models \cite{Wang2018-ti}.

    \textbf{Adversarial Imitation.}
    While extending the \textit{generative adversarial imitation learning} paradigm
    to learn imitation policies from proprioceptive state-only demonstrations
    has been previously explored \cite{Merel2017-lo},
    to the best of our knowledge only \cite{Torabi2018-nb}
    has dealt with visual state representation in this setting.
    \textit{Generative adversarial imitation from observation} \cite{Torabi2018-nb}
    reports state-of-the-art performance in state-only imitation
    against non-adversarial baselines.
    We will therefore compare our models to this baseline in \textsc{Section} \ref{resultsection}.
    Orthogonal to the control tasks considered in this work,
    Self-Attention GAN (SAGAN) \cite{Zhang2019-eb}
    reports state-of-the-art results in image generation tasks
    by making both the generator and discriminator self-attentive.
    Our method can be viewed as a bridge between SAGAN and GAIL, which we enriched by adding temporal
    relational learning capabilities by working over sequences of states,
    in addition to the spatial relational aspect.
    Overcoming GAIL's challenges such as
    sample-inefficiency \cite{Blonde2019-vc,Kostrikov2019-jo}
    and its propensity to mode collapse \cite{Hausman2017-hb,Li2017-sb}
    has been the focus of GAIL research in recent years.
    These advances are orthogonal to our approach.
    We believe our work to be the first to apply relational learning
    via non-local modeling to adversarial imitation learning.

    \textbf{Relational learning.}
    Techniques aiming to exploit the inherent structure of the state in domains or
    applications where \textit{relational inductive biases} \cite{Battaglia2018-vv}
    can be leveraged have recently gained in popularity.
    Due to their flexibility, graphs constitute the archetype structure at the root of many approaches
    explicitly looking at pairwise relationships between objects present in the state.
    Graph neural networks (GNNs) are the most common deep structures in the field.
    GNNs have been successfully used to model locomotion policies
    able to transfer across different body shapes and impairement conditions
    \cite{Wang2018-ot},
    to model interactions in physical systems \cite{Battaglia2016-xz},
    to improve inter-agent communications in multi-agent scenarios \cite{Jiang2018-go,Hoshen2017-vj},
    for gait generation \cite{Kipf2018-gs},
    for skeleton-based action recognition \cite{Shi2019-jr,Shi2019-lx}
    and
    to enhance grasp stability in dexterous robotics \cite{Garcia-Garcia2019-mp}.
    While GNNs have proven to be effective at learning relationships between objects,
    their explicitly defined structure
    cannot perform \textit{object discovery} directly from visual states in practice.
    Self-attention provides a solution as it considers the relationships between every pair
    of atomic entities in the input features (\textit{e.g.}, convolutional feature maps).
    By using a self-attention mechanism, relation networks \cite{Santoro2017-nb} learn a pairwise
    function of feature embeddings for every pair of input regions (pixels or receptive fields) to
    discover the relationships that are the most useful to solve
    question answering tasks.
    In control, self-attention has been used to play StarCraft II mini-games from raw pixel input,
    a challenging multi-agent coordination task \cite{Zambaldi2019-bd}.
    Instead, we focus on locomotion and learn limb coordination from visual input with the additional difficulty of
    having to deal with continuous action spaces and third-person side view camera state spaces
    with a dynamic background.
    Our agents architecture differs from \cite{Zambaldi2019-bd} on several points.
    We deal with spatial and temporal relationships jointly via a combination of
    convolution and self-attention, whereas \cite{Zambaldi2019-bd}
    relies on a large stack of fully-connected layers to perform non-spatial relational learning
    or alternatively,
    on a recurrent model with many parameters to perform temporal relational reasoning.
    Our method is lightweight in comparison.
    Additionally, by using non-local modeling with a skip connection,
    our model is able to attend to local regions.
    Besides providing improvements in the RL setting,
    our method yields state-of-the-art results in imitation learning as we will describe now.

\section{Results}\label{resultsection}

    \textbf{Experimental setup.}
    In every reported experiment, we use a gradient averaging
    distributed learning scheme consisting of spawning 4 parallel learners only differing
    by their random seeds, and using an average of their gradients for every learner update.
    This orchestration choice does not cause a decrease in sample-efficiency like
    traditional multi-actors massively-distributed frameworks would, which is desirable considering the
    poor sample-efficiency vanilla GAIL already suffers from.
    We repeat every experiment with 10 random seeds.
    The same seeds are used for all our experiments.
    Each learning curve is depicted with a solid
    curve corresponding to the mean episodic return across the 10 seeds,
    surrounded by an envelope of width equal to the associated standard deviation.
    We enforce this methodology to ensure that the drawn comparisons
    are fair \cite{Henderson2018-vm}.

    We report the performance of our methods and baselines
    (described later per setting)
    in locomotion environments from the \textsc{OpenAI Gym} \cite{Brockman2016-un} library
    based in the \textsc{MuJoCo} physics engine \cite{Todorov2012-gc}.
    Simulator states are rendered using the \textsc{OpenAI Gym} headless renderer
    with the default camera view for the given environment
    to ensure the reproducibility of the entirety of our experiments.
    After converting the rendered frames to gray-scale and stacking the $k$ most recent ones,
    the input state contains $84 \times 84 \times k$ pixels.

    In addition to the mean episodic return across random seeds,
    we also report the \textit{Complementary Cumulative Distribution Function} (CCDF),
    also sometimes called the \textit{survival function},
    for every experiment.
    The CCDF shows an estimate of how often the return is above a given threshold.
    In contrast with the mean, the CCDF does not taint
    the performance of the best seeds with the performance
    of the worst ones, which is an especially desirable property since we work over many seeds.

    We trained the expert policies used in this work in the environments our agents
    interact with.
    Experts are trained
    using the proprioceptive state representation
    with the PPO algorithm \cite{Schulman2017-ou}
    as it provides the best wall-clock time,
    being an on-policy model-free continuous control method.
    Expert training was set to proceed for 10 million timesteps.
    To collect demonstrations for the trained experts,
    we ran the expert in its environment while rendering every visited state
    using the same rendering scheme as with the policy
    ($84 \times 84 \times 1$ pixels per state),
    and saved these visual state representations as demonstrations.
    A demonstration is therefore a video of the expert
    interacting with its environment for an episode.
    The frames are saved with the same frame-rate as when interacting,
    which is an important property to preserve in order
    for the agent to match the expert velocity.
    For reproducibility purposes, we used the default settings of the simulator
    (\textsc{MuJoCo} version 2.0 binaries).

    Note, we used layer normalization \cite{Ba2016-bs} in the dense layers
    of every network. See supplementary for more details.


    \begin{figure}
    \centering
    \begin{subfigure}[t]{.48\linewidth}
    \includegraphics[width=\linewidth]{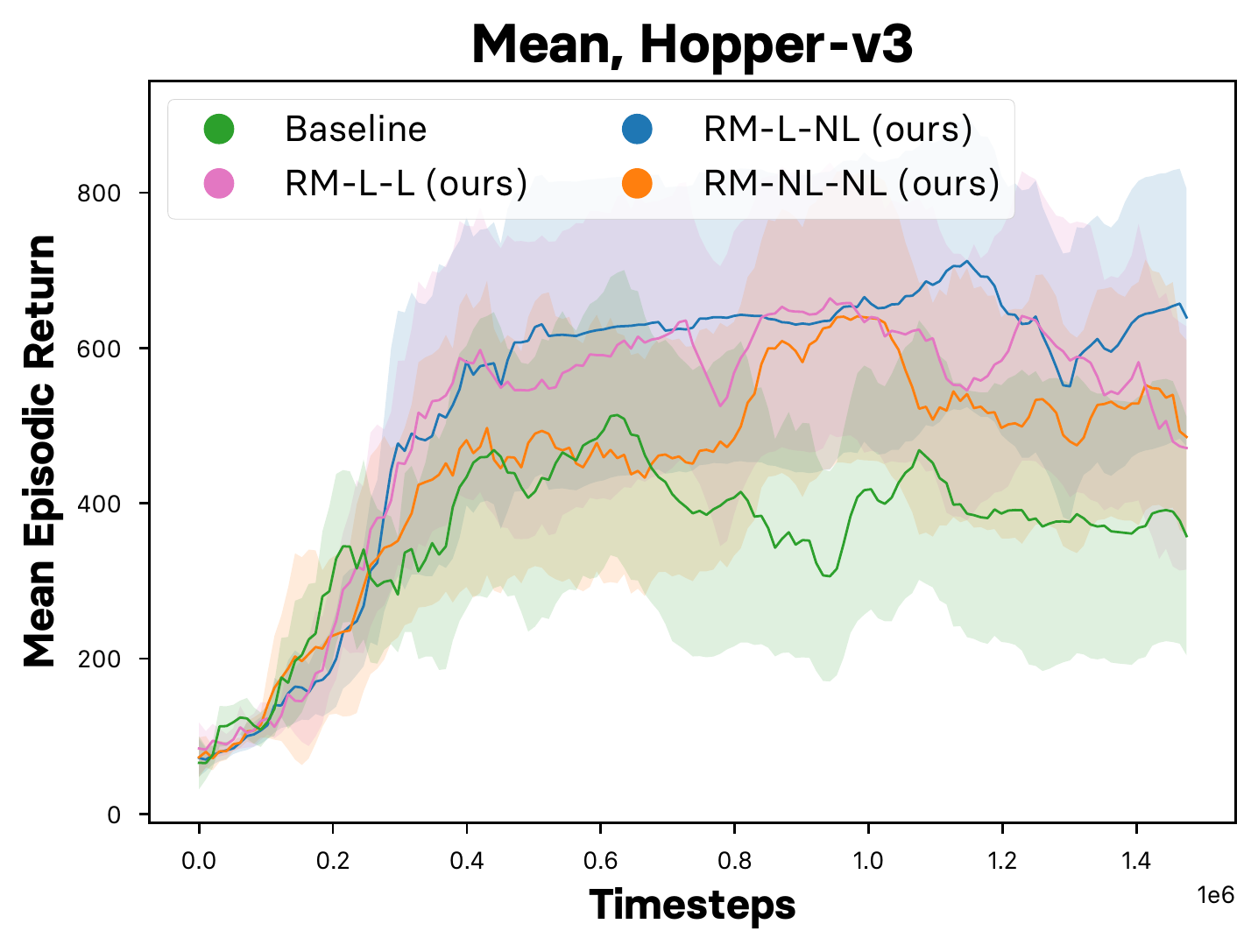}
    \end{subfigure}
    \begin{subfigure}[t]{.48\linewidth}
    \includegraphics[width=\linewidth]{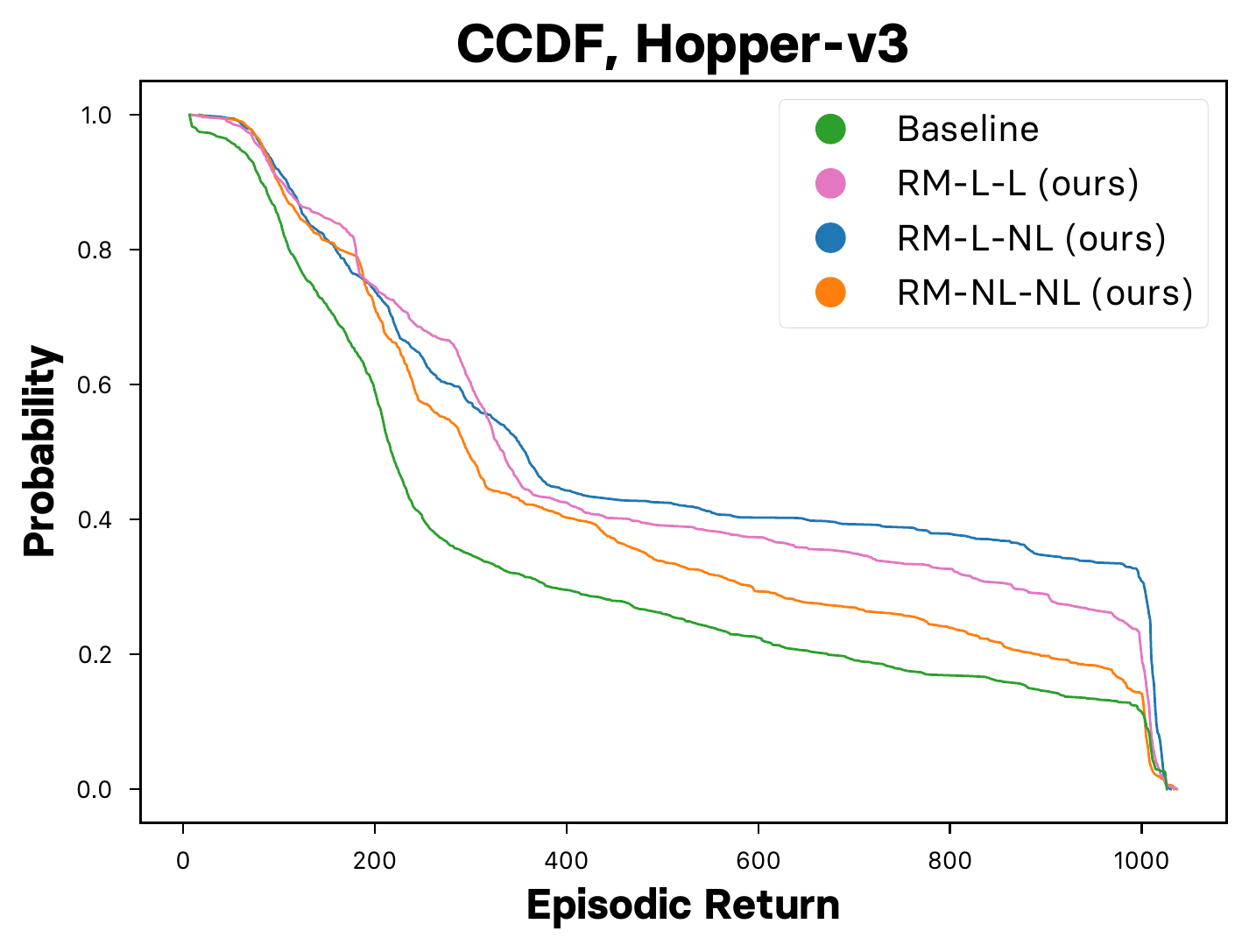}
    \end{subfigure}
    \begin{subfigure}[t]{.48\linewidth}
    \includegraphics[width=\linewidth]{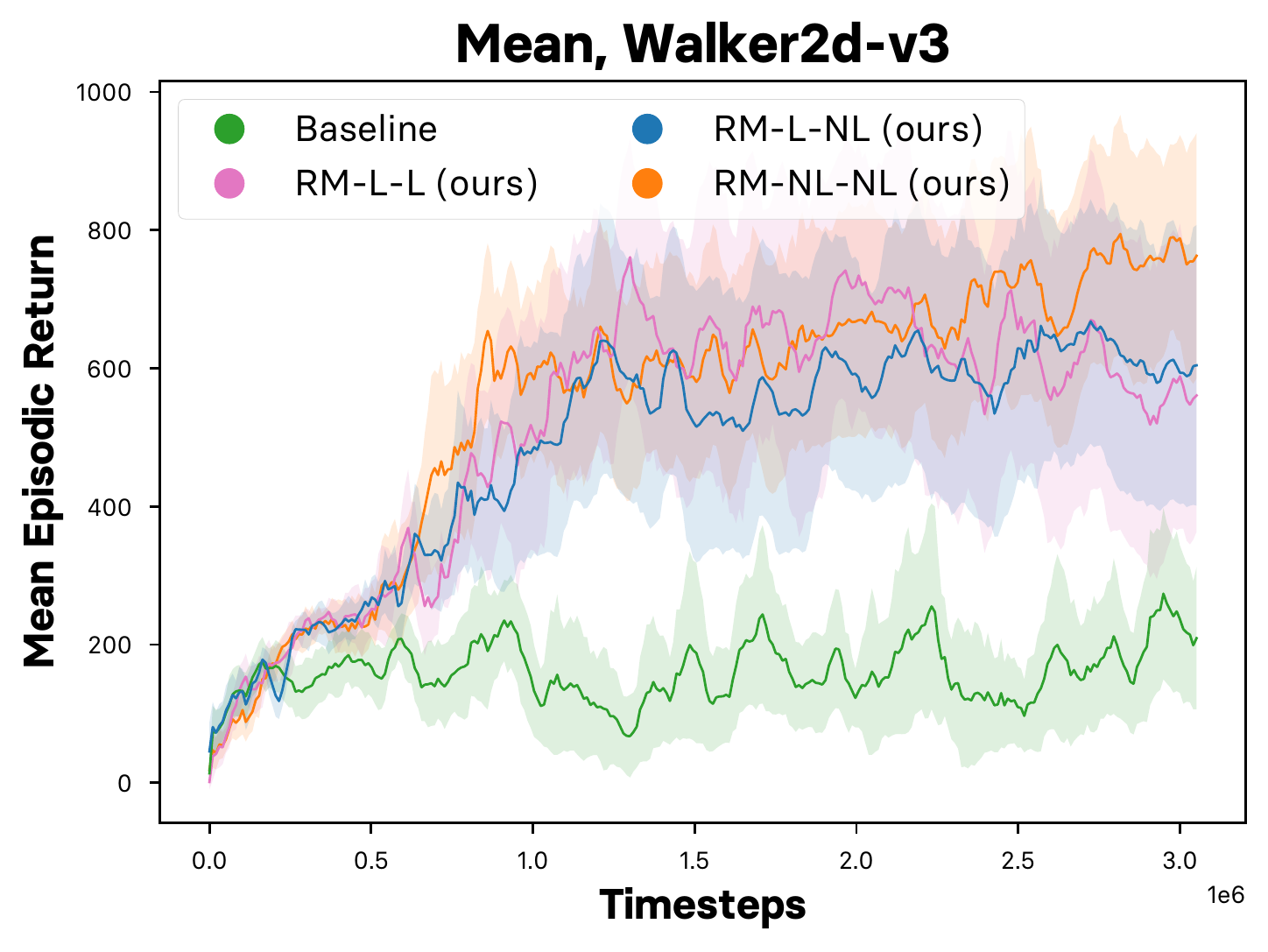}
    \end{subfigure}
    \begin{subfigure}[t]{.48\linewidth}
    \includegraphics[width=\linewidth]{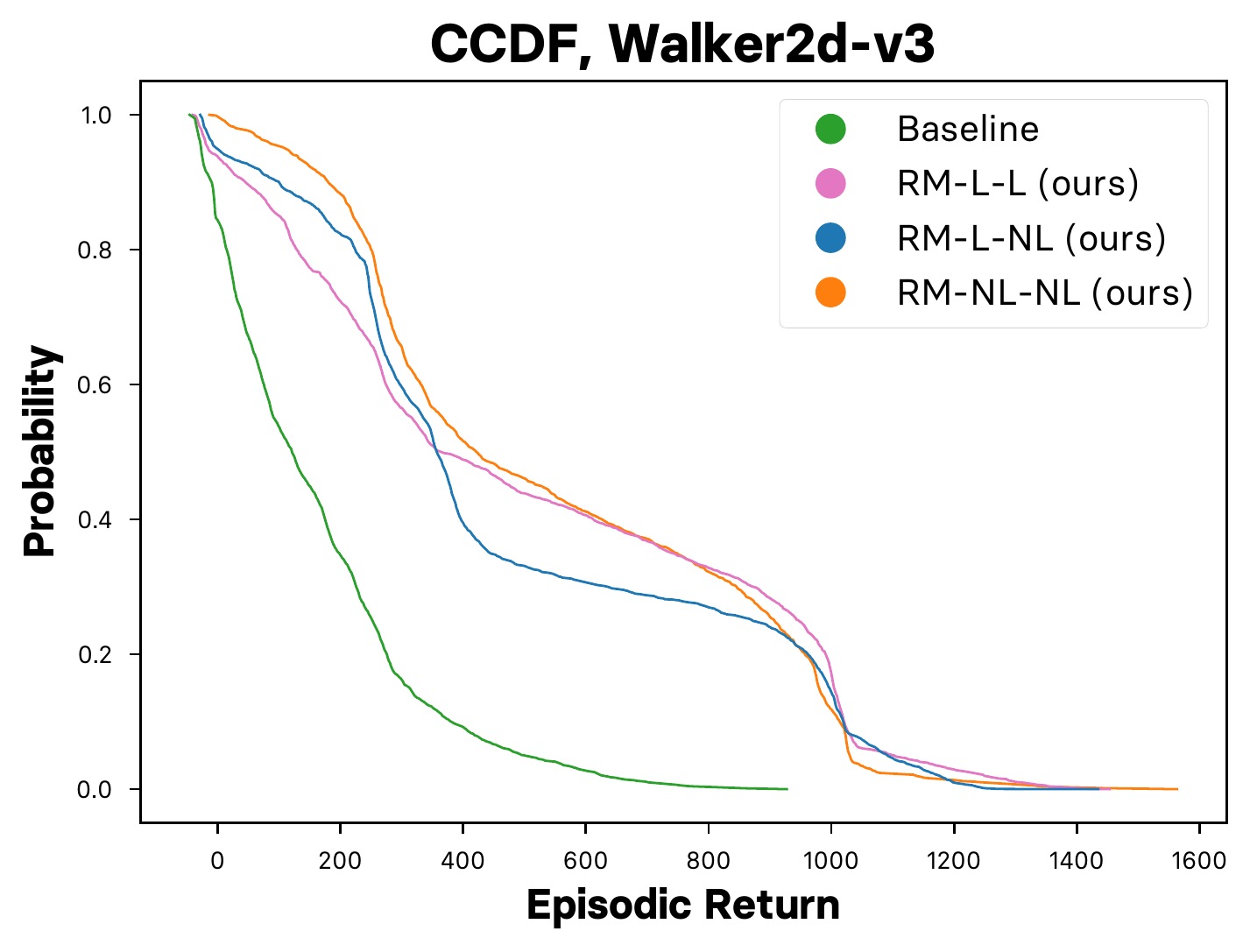}
    \end{subfigure}
    \begin{subfigure}[t]{\linewidth}
    \centering
    \begin{tabular}{l|c|c}
    \multirow{3}{*}{\textsc{Method}} & \multicolumn{2}{c}{\textit{area under CCDF for...}} \\
    & \textsc{Hopper-v3} & \textsc{Walker2d-v3} \\
    \hline\hline
    Baseline & 524,189 & 479,430 \\ \hline
    \rowcolor{Gray0}
    \textsc{RM-L-L} (ours) & 731,871 & 1,539,345 \\ \hline
    \rowcolor{Gray1}
    \textsc{RM-L-NL} (ours) & \textbf{771,666} & 1,444,834 \\ \hline
    \rowcolor{Gray1}
    \textsc{RM-NL-NL} (ours) & 645,558 & \textbf{1,640,749} \\ \hline
    \end{tabular}
    \end{subfigure}
    \caption{Imitation learning performance comparison with 8 demonstrations and $k=4$.}
    \label{fig:demoanalysis8}
    \end{figure}

    \begin{figure}
    \centering
    \begin{subfigure}[t]{.48\linewidth}
    \includegraphics[width=\linewidth]{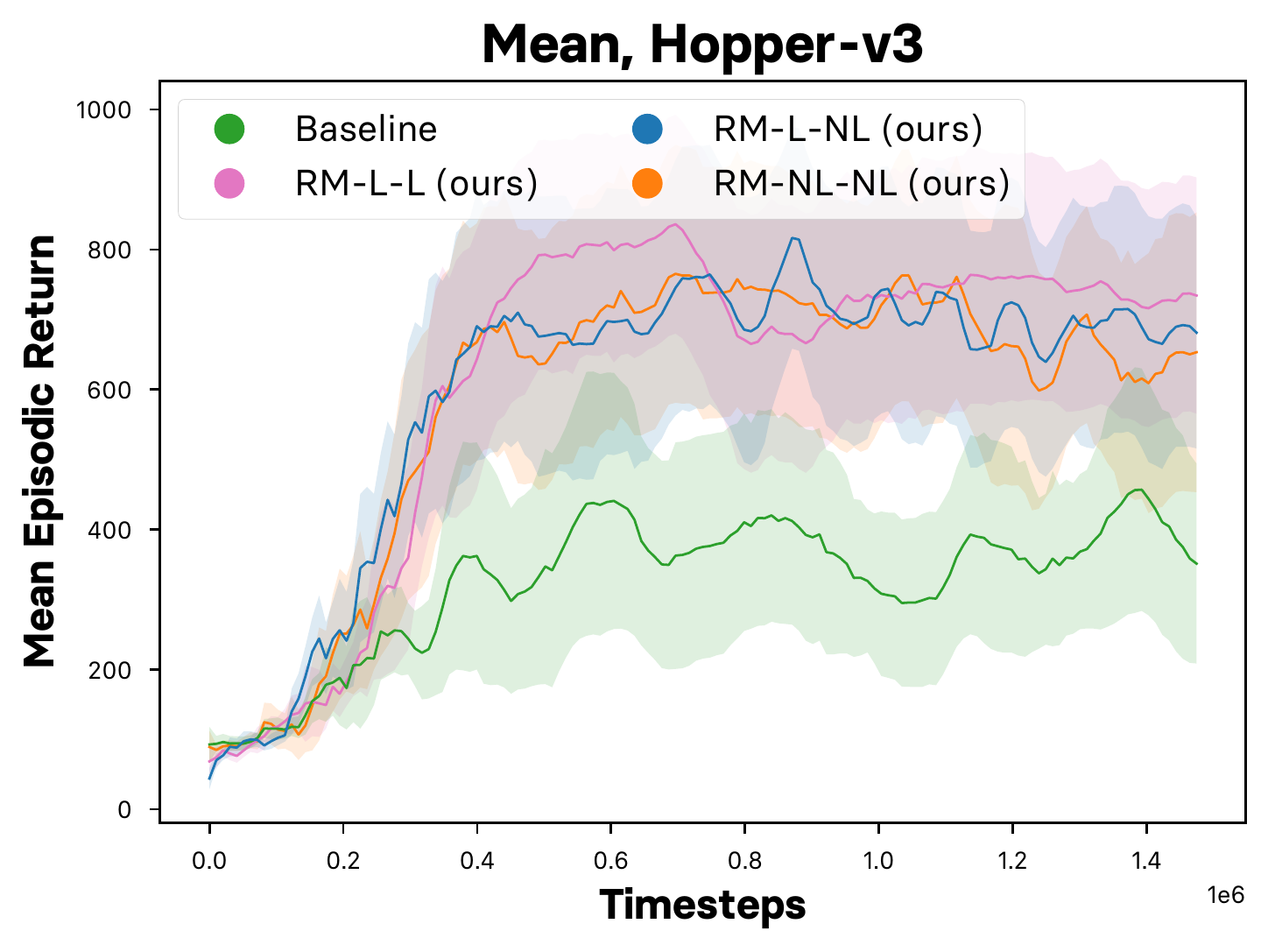}
    \end{subfigure}
    \begin{subfigure}[t]{.48\linewidth}
    \includegraphics[width=\linewidth]{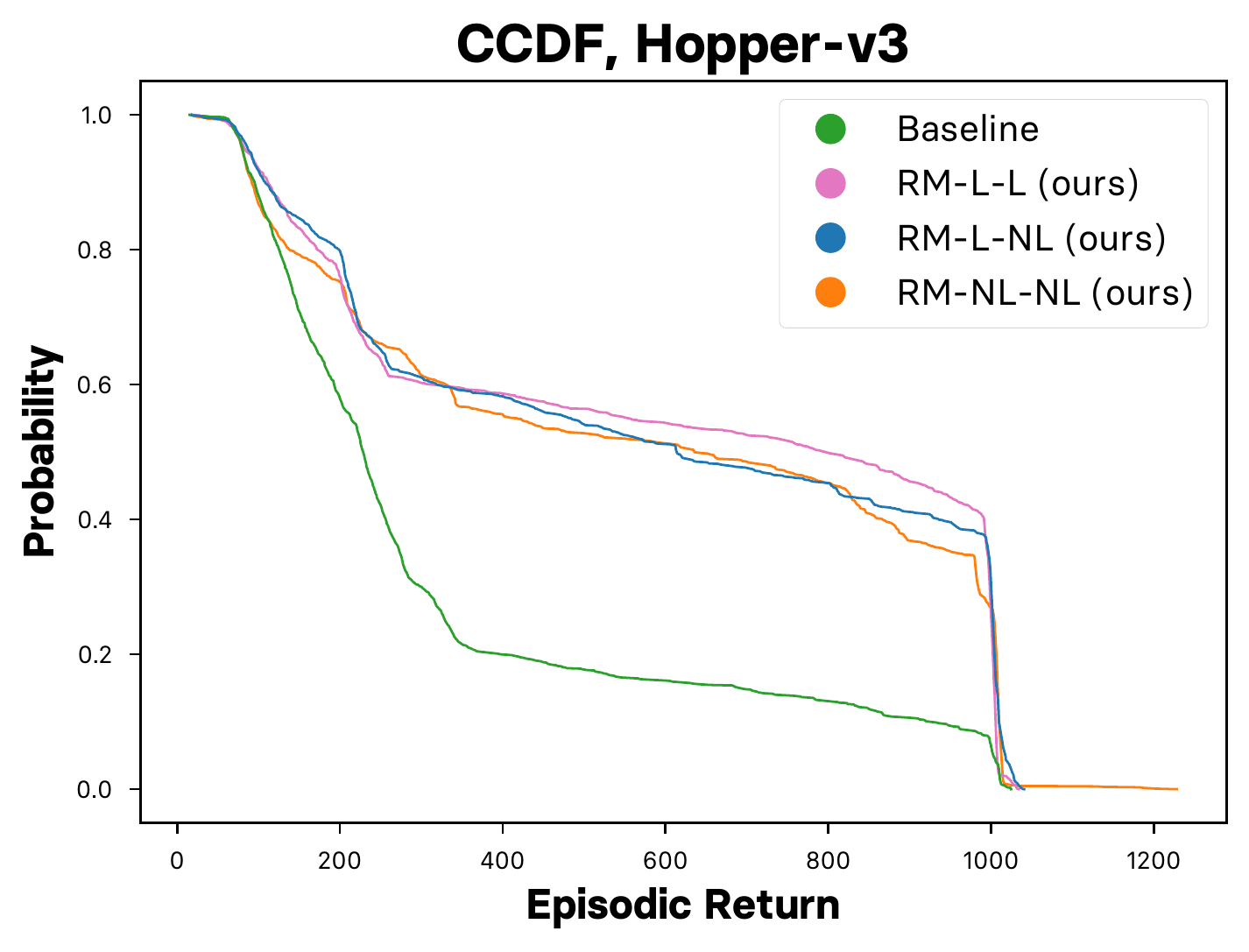}
    \end{subfigure}
    \begin{subfigure}[t]{.48\linewidth}
    \includegraphics[width=\linewidth]{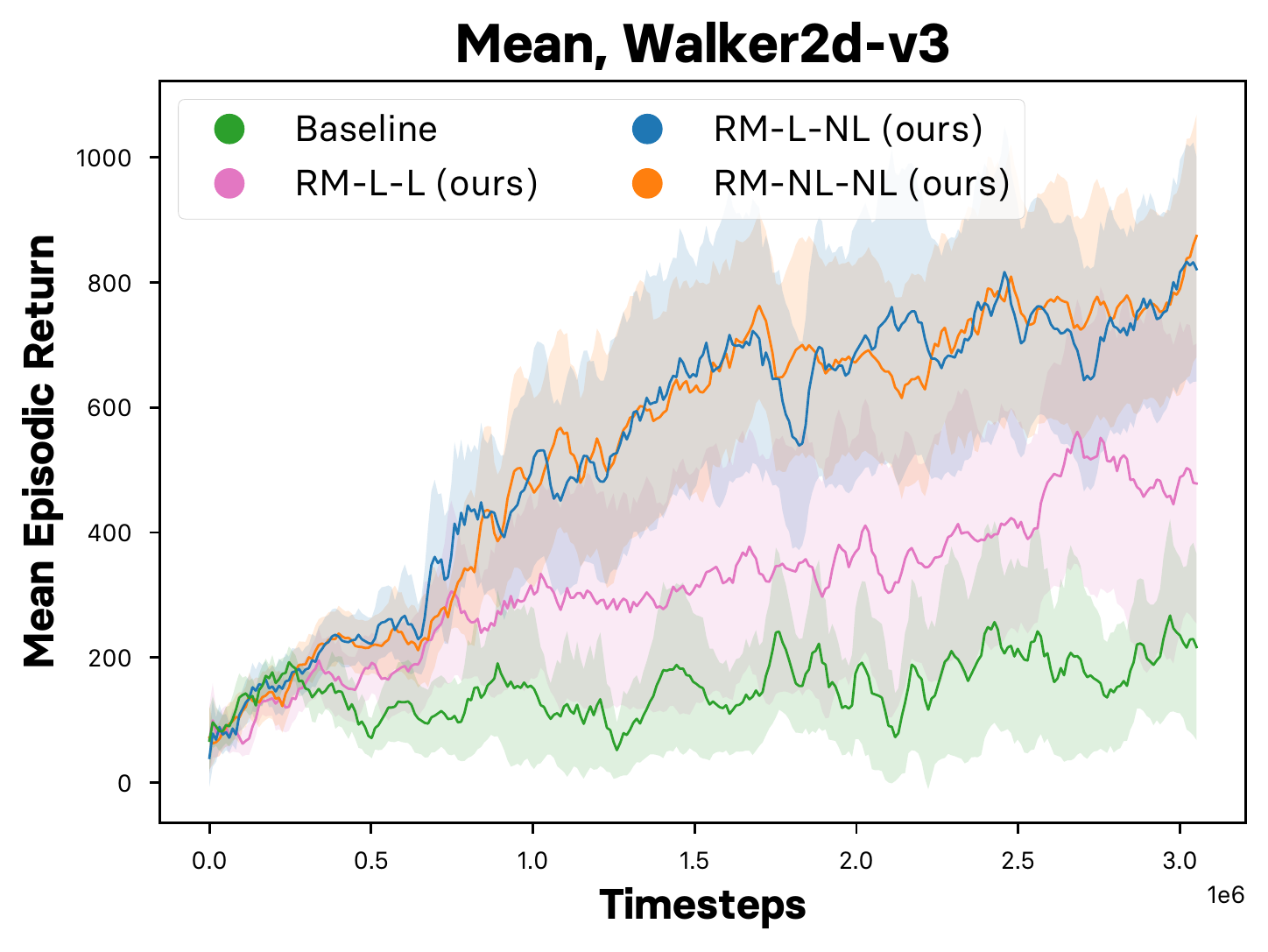}
    \end{subfigure}
    \begin{subfigure}[t]{.48\linewidth}
    \includegraphics[width=\linewidth]{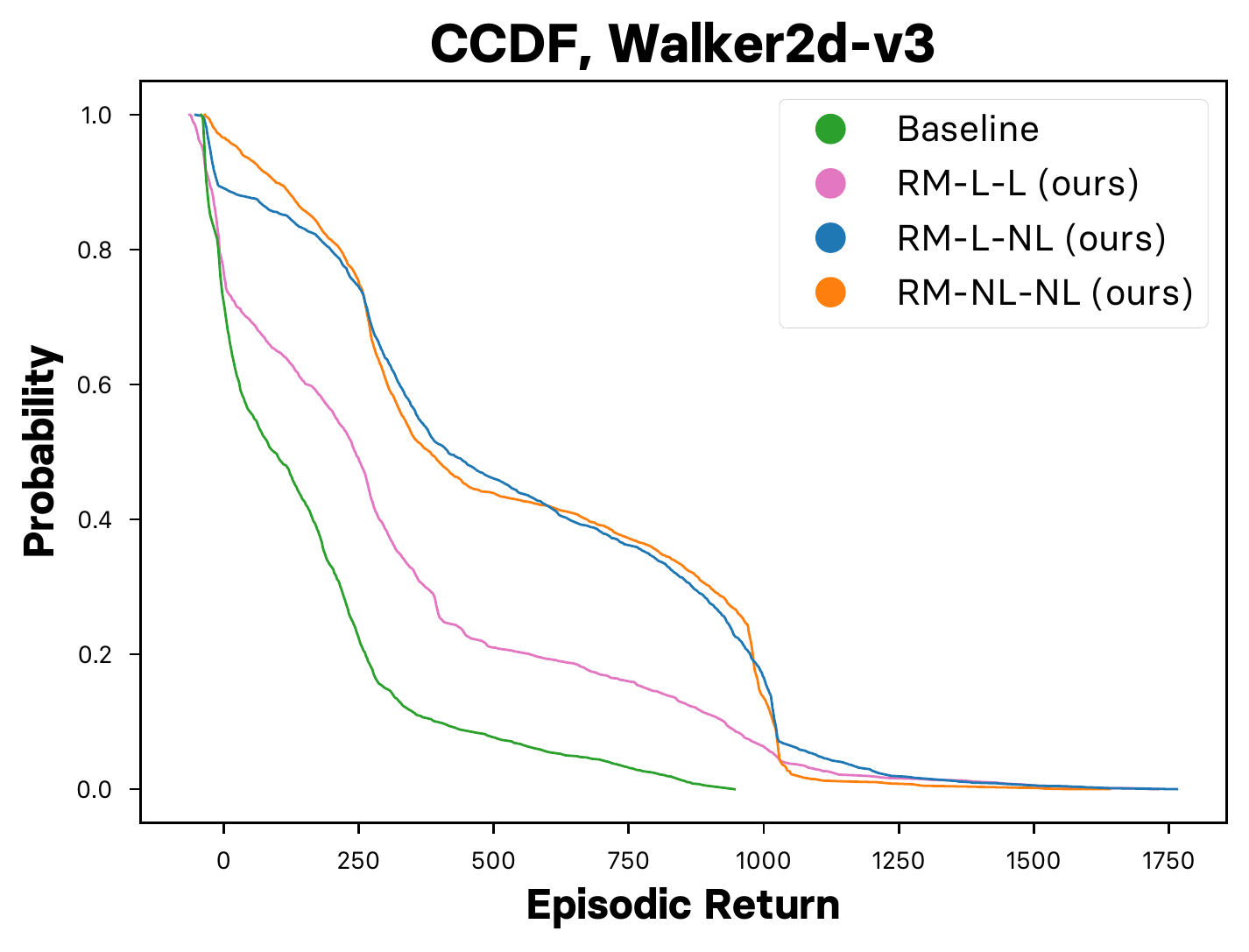}
    \end{subfigure}
    \begin{subfigure}[t]{\linewidth}
    \centering
    \begin{tabular}{l|c|c}
    \multirow{3}{*}{\textsc{Method}} & \multicolumn{2}{c}{\textit{area under CCDF for...}} \\
    & \textsc{Hopper-v3} & \textsc{Walker2d-v3} \\
    \hline\hline
    Baseline & 467,803 & 457,018 \\ \hline
    \rowcolor{Gray0}
    \textsc{RM-L-L} (ours) & \textbf{892,730} & 846,887 \\ \hline
    \rowcolor{Gray1}
    \textsc{RM-L-NL} (ours) & 866,267 & \textbf{1,604,128} \\ \hline
    \rowcolor{Gray1}
    \textsc{RM-NL-NL} (ours) & 846,395 & \textbf{1,603,569} \\ \hline
    \end{tabular}
    \end{subfigure}
    \caption{Imitation learning performance comparison with 8 demonstrations and $k=8$.}
    \label{fig:longerhistory}
    \end{figure}

    \textbf{Imitation learning results.}
    In \textsc{Figures} \ref{fig:demoanalysis8} and \ref{fig:longerhistory},
    we compare the performance
    of three different network configurations of \textsc{RM}
    --- \textsc{RM-L-L} (\textit{local policy, local value}),
    \textsc{RM-L-NL} (\textit{local policy, non-local value}),
    and \textsc{RM-NL-NL} (\textit{non-local policy, non-local value}) ---
    against the \textit{baseline}.
    The \textsc{RM} configurations only differ by their use of relational blocks in
    the reward, policy and value modules, and are summarized in \textsc{Table} \ref{fig:yn}.
    The baseline closely resembles \cite{Torabi2018-nb},
    the variant of GAIL \cite{Ho2016-bv}
    reporting SOTA performance in imitation from demonstrations without actions.
    The only differences from \cite{Torabi2018-nb} are that we modulate the number of stacked
    frames in the visual input state, and that we adopt network architectures that
    make the comparisons against \textsc{RM} fair.
    The baseline corresponds to \textsc{RM} without relational modules,
    as summarized in \textsc{Table} \ref{fig:yn}.
    Note, \cite{Torabi2018-nb} only reports results for environments less complex that Walker2d
    in the studied setting.
    \begin{table}
    \centering
    \begin{tabular}{l|c|c|c}
    \textit{Non-local?} & reward & policy & value \\
    \hline\hline
    Baseline & no & no & no \\ \hline
    \rowcolor{Gray0}
    \textsc{RM-L-L} (ours) & \textbf{yes} & no & no \\ \hline
    \rowcolor{Gray1}
    \textsc{RM-L-NL} (ours) & \textbf{yes} & no & \textbf{yes} \\ \hline
    \rowcolor{Gray1}
    \textsc{RM-NL-NL} (ours) & \textbf{yes} & \textbf{yes} & \textbf{yes} \\ \hline
    \end{tabular}
    \caption{Use of relational blocks in the different modules.}
    \label{fig:yn}
    \end{table}
    In \textsc{Figure} \ref{fig:demoanalysis8},
    we observe that:
    a) non-local modeling has the most significant effect when used in the reward module compared to other modules,
    b) all the \textsc{RM} variants perform similarly, and
    c) the baseline does not take off in the more complex Walker2d environment, unlike \textsc{RM}.

    Additionally, \textsc{Figure} \ref{fig:longerhistory} shows that,
    by increasing the input sequence length $k$ from $4$ to $8$,
    \textsc{RM} achieves better performance in the Hopper environment, with a +21\% increase for all methods.
    In the Walker2d environment however,
    while \textsc{RM-L-NL} benefits from a +11\% increase in performance,
    \textsc{RM-L-L} suffers from a -45\% decrease,
    but still scores well above the baseline.
    This shows that using relational learning in the value and (or) policy can help dealing
    with longer input sequences, a particularly valuable observation for POMDPs that require such
    long input history to alleviate poor state observability.
    Finally, the baseline suffers from a -7\% performance drop in every environment when increasing $k$
    from 4 to 8, further widening the gap with \textsc{RM}.

    \begin{figure}
    \centering
    \begin{subfigure}[t]{.48\linewidth}
    \includegraphics[width=\linewidth]{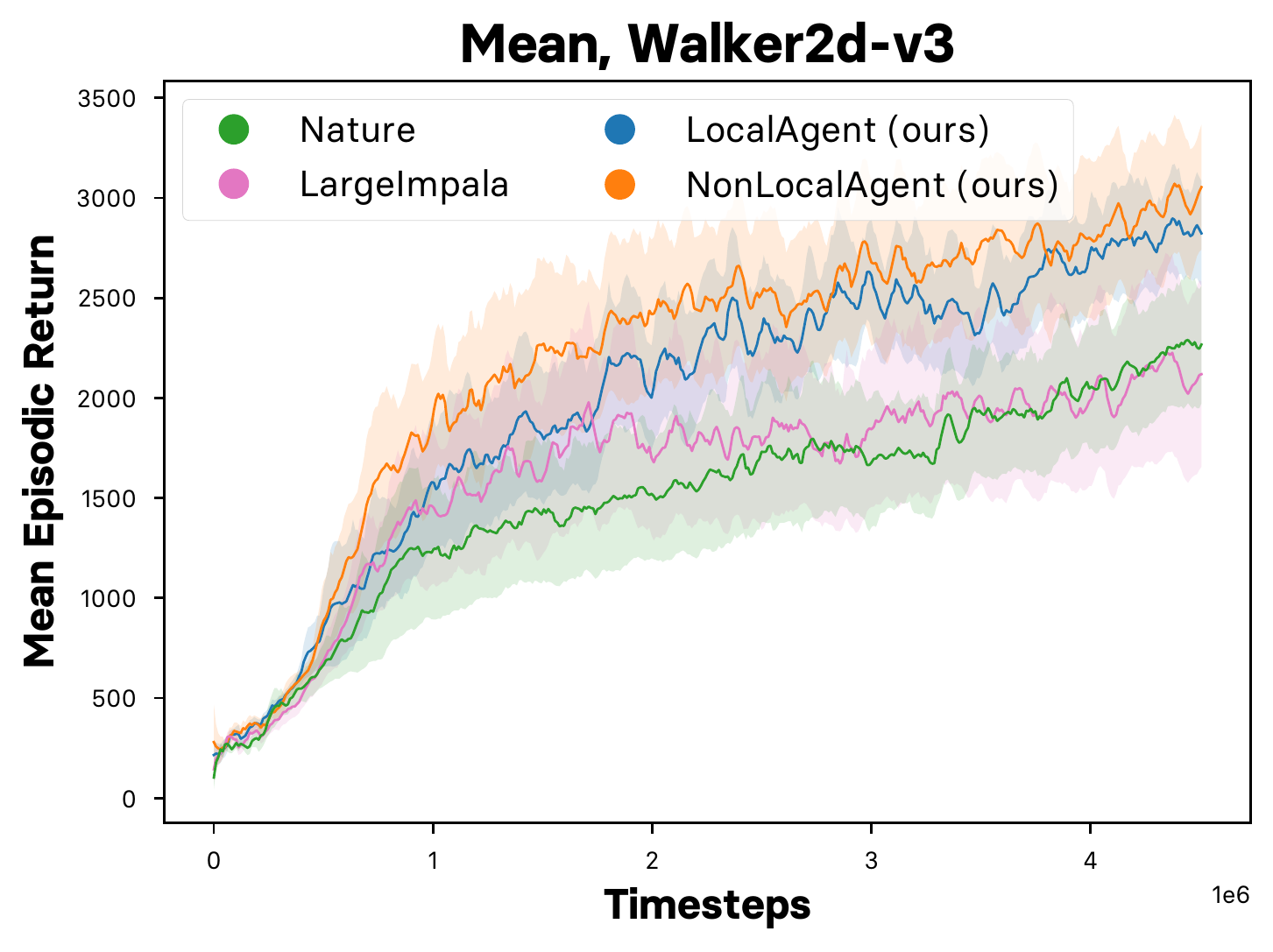}
    \end{subfigure}
    \begin{subfigure}[t]{.48\linewidth}
    \includegraphics[width=\linewidth]{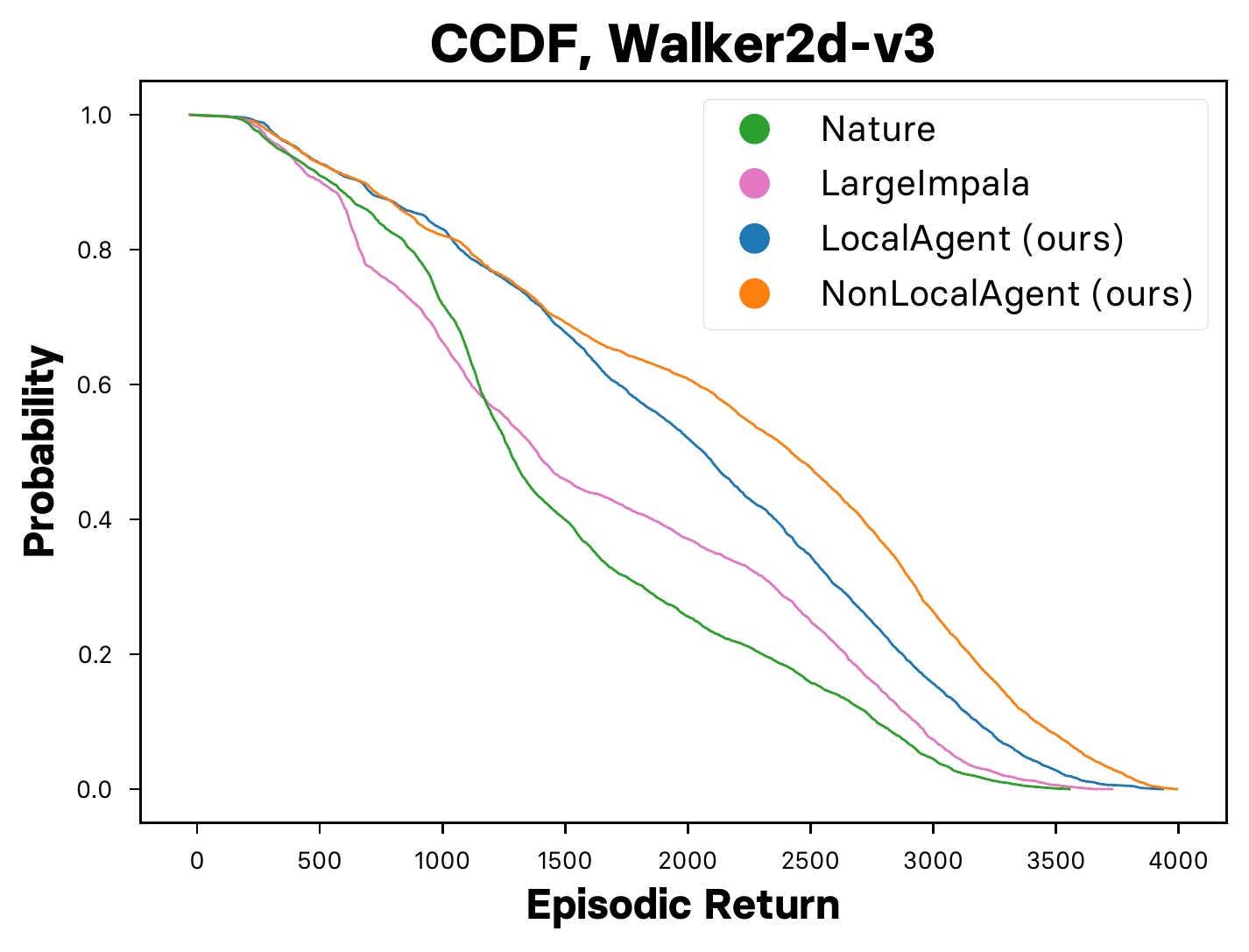}
    \end{subfigure}
    \begin{subfigure}[t]{\linewidth}
    \centering
    \begin{tabular}{l|c}
    \multirow{3}{*}{\textsc{Method}} & \textit{area under CCDF for...} \\
    & \textsc{Walker2d-v3} \\
    \hline\hline
    \textsc{Nature} \cite{Mnih2015-iy} & 6,582,843 \\ \hline
    \textsc{LargeImpala} \cite{Espeholt2018-gl} & 7,112,912 \\ \hline
    \rowcolor{Gray0}
    \textsc{LocalAgent} (ours) & 8,774,574 \\ \hline
    \rowcolor{Gray1}
    \textsc{NonLocalAgent} (ours) & \textbf{9,652,209} \\ \hline
    \end{tabular}
    \end{subfigure}
    \caption{RL performance comparison with $k=4$.}
    \label{fig:rl}
    \end{figure}

    \textbf{Reinforcement learning results.}
    In \textsc{Figure} \ref{fig:rl}, we compare the performance
    of several architectures, previously described in \textsc{Table} \ref{fig:archicomparison}.
    We train these by RL, with PPO \cite{Schulman2017-ou},
    using the reward from the environment.
    The results show that the \textsc{LocalAgent} outperforms the
    \textsc{Nature} \cite{Mnih2015-iy} and \textsc{LargeImpala} \cite{Espeholt2018-gl} baselines.
    The performance is further increased by the \textsc{NonLocalAgent},
    using relational modules in both the policy and value ($\sim$10\% increase).

\section{Conclusion}
    In this work, we introduced \textsc{RM}, a new method for visual imitation learning from observations
    based on GAIL \cite{Ho2016-bv},
    that we enriched with
    the capability to consider spatial and temporal long-range relationships in the input,
    allowing our agents to perform relational learning.
    Since the significant gains in sample-efficiency and overall performance enabled by our method
    stem from an architecture enrichment, \textsc{RM} can be directly combined with methods
    addressing GAIL sample-inefficiency by algorithmic enhancements.
    The obtained results are in line with our initial conjecture about the usefulness of self-attention
    to solve locomotion tasks.
    Our method is able to work with high-dimensional state spaces, such as video sequences,
    and shows resilience to periodic limb obstruction on the pixel input and video demonstration misalignment.
    Finally, we show the effect of self-attention in the different components of our model
    and show outcomes on policy improvement,
    policy evaluation, and reward learning.
    The most significant impact was observed when we used self-attention for reward learning.

\section{Future Work}
    In future work, an investigation of visual relational learning could help agents to
    better cope with induced simulated body impairments \cite{Wang2018-ot} in locomotion tasks
    and predict the impact of proposed changes on the ensuing walking gait \cite{Lee2019-yc}
    when working solely with visual state representations.
    Another avenue of improvement could be to leverage other modalities relevant for skeleton-based
    locomotion (\textit{e.g.}, limb morphology, kinematics) to solve the upstream task of learning
    an accurate \textit{inverse dynamics} model
    \cite{Hanna2017-iz,Torabi2018-fg,Ganin2018-jb}.
    Using this model, one could then learn a mimic on the the richer, action-augmented demonstrations.

\clearpage

{\small
\bibliographystyle{ieee}
\bibliography{bibliography}
}

\end{document}